\documentclass[
   onecolumn
]{article}				         

\usepackage[
   english                    
]{babel}
\usepackage[
   babel,
   german=quotes
]{csquotes}                   
\usepackage[
   T1                         
]{fontenc}
\usepackage[
   a4paper,                   
   top=3cm,
   bottom=4cm,                
   outer=2.5cm,
   inner=3cm,                 
]{geometry}
\usepackage[
  backend=biber,
  style=numeric,              
  sorting=none                
]{biblatex}
\usepackage{url}              
\usepackage[noblocks]{authblk}
\usepackage{lipsum}           
\usepackage{multicol}         
\usepackage{sectsty}          
\usepackage{amsmath}          
\usepackage{amssymb}          
\usepackage{hhline}           
\usepackage{float}            
\usepackage[font=small,labelfont=bf]{caption}
\usepackage{graphicx}         
\usepackage{subfig}           
\usepackage{svg}              
\usepackage[hidelinks]{hyperref} 
\usepackage[
    type={CC},
    modifier={by-sa},
    version={4.0},
    imagemodifier={-80x15},
]{doclicense}                 

\begin{document}


\title{Tabular Learning: Encoding for Entity and Context Embeddings}
\author{Fredy Reusser}
\affil{Institute for Data Applications and Security, School of Engineering and Computer Science,
      Bern University of Applied Sciences, Biel, Switzerland
      }
\maketitle

\doclicenseThis

\renewenvironment{abstract}
{
   \hspace{-0.55cm}
   \vspace{0.2cm}
   \rule{\linewidth}{.5pt}
   \textbf{\abstractname}
   \list{}{
      \setlength{\leftmargin}{0mm}
      \setlength{\rightmargin}{\leftmargin}
   }
   \item\relax
}
{
   \endlist
   \vspace{-0.4cm}
   \rule{\linewidth}{.5pt}
   \vspace{0cm}
}

\begin{abstract}

   Examining the effect of different encoding techniques on entity and context embeddings,
   the goal of this work is to challenge commonly used Ordinal encoding for tabular learning.
   Applying different preprocessing methods
   and network architectures over several datasets resulted in a benchmark on how the encoders
   influence the learning outcome of the networks. By keeping the test, validation and training data
   consistent, results have shown that ordinal encoding is not the most suited encoder for categorical
   data in terms of preprocessing the data and thereafter, classifying the target variable correctly.
   A better outcome was achieved, encoding the features based on string similarities by computing a
   similarity matrix as input for the network. This is the case for both, entity and context embeddings,
   where the transformer architecture showed improved performance for Ordinal and Similarity encoding
   with regard to multi-label classification tasks.     
\end{abstract}


\renewcommand{\thesection}{\Roman{section}}
\sectionfont{\Large}
\subsectionfont{\large}

\newenvironment{Table}
  {\par\bigskip\noindent\minipage{\columnwidth}\centering}
  {\endminipage\par\bigskip}

\newenvironment{Figure}
  {\par\medskip\noindent\minipage{\linewidth}\centering}
  {\endminipage\par\medskip}

\begin{multicols}{2}
   \section{Introduction}
   \noindent
   Tree based methods such as Random Forests\cite{Breiman2001Oct} and gradient boosted ensemble
   methods (e.g. XGBoost)\cite{Chen2016Aug} are popular machine learning models which are used for a
   wide range of different applications. Being capable of handling homogenous (solely continuous or
   discrete predictors) as well as heterogenous tabular data, these models are appreciated
   for their ease of use. It is not surprising that the mentioned algorithm family still dominates
   the structured data domain, as recent benchmarks have shown\cite{Grinsztajn2022Dec}\cite{Shwartz-Ziv2021Jun}.\par\noindent
   Though neural networks are still being outperformed, continued adaption of methods and architectures
   from e.g. the field of text analysis yielded interesting results within the past several years.
   The use of embeddings for categorical data originally proposed by Cheng Guo and Felix Berkham
   \cite{Guo2016Apr} opened up a novel way of handling categorical data. Over the course of time the
   outcomes were new deep learning architectures such as e.g. TabNet\cite{Arik2019Aug},
   TabTransformer\cite{Huang2020Dec} and TabPFN\cite{Hollmann2022Jul}.\\
   \noindent
   The following work is focusing not on the further development of a potential architecture but rather
   on the data input of such a model. Considering the survey on categorical data for neural
   networks\cite{Hancock2020Dec}, the goal is to determine the effect of different encoding techniques
   on the entity and context model respectively. Therefore, discretization of continuous features
   present in the used datasets will be applied as a first step. To evaluate potential differences between several
   encoding methods, the model architecture from Guo \& Berkham as well as the TabTransformer are implemented during a second
   step to obtain entity and context embeddings as well as the model predictions.
      
   \section{Discretization} \label{discretization}
   \noindent
   Discretization describes the procedure of splitting continuous variables and assigning their values to 
   a number of bins in order to get a set of categories. Methods used to obtain said output can
   be assigned to two sets of characteristics: unsupervised/supervised and global/local\cite{gupta2010clustering}.
   Unsupervised methods aim at learning internal data patterns by processing large amounts of data, whereas
   supervised methods utilize a predefined target variable in order to search for not yet known data structures\cite{sathya2013comparison}.
   On the other hand, it can be distinguished between algorithms who operate on a global (considering all input
   features at once before decision-making) or local scale, processing one variable at the time\cite{Chmielewski1996Nov}.
   Examples for an unsupervised-global method is k-means\cite{Likas2003Feb} and for the
   supervised-local variant a decision tree model (when performed as entropy minimization method on a single
   input feature)\cite{kotsiantis2006discretization}. Fundamental approaches like the equal-interval-width\cite{chan1991determination}
   or the equal-frequency-per-interval\cite{wong1987synthesizing} method are not considered as a preprocessing step for the experimental setup.

   \subsection*{K-means}
   \noindent
   K-means aims at dividing a dataset into $K$ partitions by minimizing the sum of Euclidean
   distances between the data points $x_i$ and their corresponding centroid $c_j$ within cluster $S_j$\cite{macqueen1967some}.
   \begin{equation}
      \label{eq:1}
      J = \sum_{j=1}^{K} \sum_{i=1}^{N} ||x_i^{(j)}-c_j||^2
   \end{equation}
   \noindent
   Optimized by minimizing the intra cluster variance as shown in equation \ref{eq:2}, Lloyds algorithm (k-means) proceeds to assign a new data point
   to a possible cluster which lets the cluster variance grow the least per iteration done\cite{lloyd1982least}.
   \begin{equation}
      \begin{split}
         \label{eq:2}
         S_j^{(t)} \supset \{x_i: ||x_i^{(j)}-c_{j^*}||^2 \leq ||x_i^{(j)}-c_{j}||^2\}\\
         \text{for}\; j \neq j^*, j = 1,...,k 
      \end{split}
   \end{equation}
   \noindent
   After assigning each data point $x_i$ to a cluster $S_j$, the centroids $c_j$ need to be updated as shown in eq. \ref{eq:3}
   in order to represent the cluster-center during the next iteration.
   \begin{equation}
      \label{eq:3}
      c_j^{(t+1)} = \frac{1}{|S_j^{(t)}|} \sum_{x_j \in S_j^{(t)}} x_i
   \end{equation}
   Although the procedure of k-means is very powerful, it comes with a major disadvantage with regard
   to its starting criterion. Usually the right number of partitions $K$ is unknown but needs to be
   selected by the user in advance. To discretize features, the method therefore demands prior knowledge
   on how many categories are appropriate. Furthermore, since the approach utilizes all inputs to calculate
   a group membership for a single data point, a sample will be affiliated over the row and not column wise as
   intended. Thus, the behaviour will not effectively result in discretization of single features.
   
   \subsection*{Decision Tree}
   Based on its predecessor ID3 (Iterative Dichotomiser 3)\cite{quinlan1986induction}, C4.5 is known as one
   of the first algorithms to build decision trees from continuous and discrete features\cite{quinlan2014c4}.
   A Decision Tree model aims to subset the feature space $X$ by hierarchical mutual exclusion, resulting
   in a set of classes $C=\{C_1,C_2,...,C_k\}$. This is done by obtaining regions $\{R_1,R_2,...,R_k\}$
   which are split by the decision node's threshold $t$.
   \begin{equation}
      \label{eq:4}
      R_1 = \{x \in \mathbb{R} | x_i > t\} \ \textrm{and} \ R_2 = \{x \in \mathbb{R} | x_i \leq t\}
   \end{equation}
   Referring to eq. \ref{eq:4}, threshold $t$ is chosen based on some impurity measure (e.g. cross-entropy)\cite{quinlan2014c4}
   to maximize the node's purity. The split should therefore produce two regions which include as many samples $x_i$
   of one class $C_k$ as possible (eq. \ref{eq:5}).
   \begin{equation}
      \label{eq:5}
      D_m(T) = -\sum_{k=1}^{K} \hat{p}_{mk} \log{\hat{p}_{mk}}
   \end{equation}
   Tree models tend to overfit extensively the more complex they grow. This behaviour leads to malperformance on
   unseen data. Considering the variance bias tradeoff, growing a less complex model which can still capture
   important data structures is desired. To achieve such a tree, cost complexity pruning\cite{breiman2017classification}
   provides an elegant way for regularization of the algorithm.
   \begin{equation}
      \label{eq:6}
      R_\alpha(T) = \sum_{m=1}^{|T|} N_mQ_m(T) + \alpha|T|,\ \alpha > 0
   \end{equation}
   By growing a very large tree $T_0$, cost complexity (weakest link) pruning searches for a subtree $T \subset T_0$
   which minimizes the cost complexity function \ref{eq:6}. While $Q_m(T)$ represents an arbitrary impurity measure,
   choosing a high $\alpha$ results in more regularization and therefore in a tree with less terminal and decision
   nodes.\\
   As a supervised approach, decision trees rely on a target variable in order to partition the feature space.
   Therefore, it is possible to create a monotonic relationship between each predictor and the target, which will
   be advantageous during model training\cite{hu2011rank}. Furthermore, not having to define a starting criterion gives the
   algorithm more flexibility in forming meaningful categories for each feature. Lastly, by preventing complexity through
   pruning, preserving relevant decision nodes during discretization will keep the number of categories within
   each feature at an acceptable occurrence rate. 

   \section{Encoders}
   \noindent
   Used to translate categorical features and represent them as numerical values, encoders play a crucial 
   preprocessing role when obtaining a suitable input format for numerical algorithms. In general, encoding methods
   can be summarized into the three groups determined, automatic and algorithmic\cite{Hancock2020Dec}.
   It is not unusual that blends between those groups are created where e.g. determined encoded variables
   are fed into automatic or algorithmic approaches. Within the determined domain, a further distinction between
   methods can be made\cite{fitkov2012evaluating}. Based on the work of Fitkov et al., an incomplete list of
   encoders and their potential affiliation to an encoder-family was constructed. Table \ref{table:1} shows all
   evaluated encoders where $k$ represents the number of classes, $n$ the base of BaseN encoding and $q$ the number
   of quantiles used for Summary encoding.
   
   \subsection*{(Rare-)Label/Ordinal Encoder}
   The widely used procedure transforms categorical variables into a discrete numerical representation.
   Known for its simplicity, the method introduces an ordinal structure which implies ordering and equal
   distances between each class $C_k$\cite{takayama2019encoding}.
   Rarelabel encoding differentiates by reducing the variable cardinality $C_k$ if a given Class frequency 
   $F_c$ falls below a predefined threshold $t$.

   \begin{equation}
      \label{eq:7}
      \begin{split}
         F_c = \frac{\text{\# of samples $x_i$ in class $C_k$}}{\text{\# of all samples $x_i$}},\\
         \text{where}\; F_c \leq t\; \text{and}\; t \in [0,1]
      \end{split}
   \end{equation}

   \subsection*{One-Hot Encoder}
   Another well known transformation method is the One-Hot encoding technique. By converting a variable with
   a set of $C$ distinct classes into a set of $X$ binary features (where $C$ and $X$ are holding an equal number of elements),
   the encoder lets the feature space grow rapidly and tends to introduce sparsity when processing
   predictors\cite{ul2019categorical}.

\end{multicols}

\noindent
\begin{Table}
\vspace{0cm}
\begin{tabular}{|p{1.4cm}|p{3.2cm}|p{1.3cm}|p{2.5cm}|p{2.3cm}|p{2.3cm}| }
   \hline
   \multicolumn{6}{|c|}{Encoder}                                                 \\
   \hline
   Family      & Name               & Acronym   & Max Feature\newline Dimensionality   & Feature-\newline Space & Class-Semantic \\
   \hhline{|=|=|=|=|=|=|}
   Index       & Label              & LE        & \hfil $1$                & Preservation & Preservation    \\
               & Rarelabel          & RE        & \hfil $1$                & Preservation & Decomposition   \\
               & Ordinal            & OE        & \hfil $1$                & Preservation & Preservation    \\
               & BaseN              & BNE       & \hfil $n$                & Expansion    & Decomposition   \\
   \hline
   Bit         & One-Hot            & OHE       & \hfil $k$                & Expansion    & Decomposition   \\
               & Binary             & BE        & \hfil $\lceil log_2{(k)} \rceil$ & Expansion    & Decomposition   \\
   \hline            
   Target      & Target (Mean)      & TE        & \hfil $1$                & Preservation & Preservation    \\
               & Leave-One-Out      & LOOE      & \hfil $1$                & Preservation & Preservation    \\
               & Weight of Evidence & WEE       & \hfil $1$                & Preservation & Preservation    \\
               & James-Stein        & JSE       & \hfil $1$                & Preservation & Preservation    \\
               & Summary (Quantile) & SE        & \hfil $q$                & Expansion    & Decomposition   \\
   \hline
   Contrast    & Backward Difference& BDE       & \hfil $k-1$              & Expansion    & Decomposition   \\
               & Helmert            & HE        & \hfil $k-1$              & Expansion    & Decomposition   \\
               & Effect (Sum)       & EE        & \hfil $k-1$              & Expansion    & Decomposition   \\
   \hline
   Others      & Frequency          & FE        & \hfil $1$                & Preservation & Preservation    \\
               & String Similarity  & STSE      & \hfil $k$                & Expansion    & Preservation    \\
   \hline
  \end{tabular}
  \captionof{table}{Encoders grouped by their characteristics}
  \label{table:1}
\end{Table}
  
\begin{multicols}{2}

   \subsection*{Target Encoder}
   Also known as Mean encoding, the method takes the target variable into account and therefore utilizes prior
   knowledge to calculate the impact a class $C_k$ could have on the target $Y$\cite{micci2001preprocessing}.
   As shown in eq. \ref{eq:8}, in the simplest case (if all classes are sufficiently large) the ratio between
   the observations $x_i$ where $Y=1$ and the overall number of observations $n$ for each class will be computed. 
   \begin{equation}
      \label{eq:8}
      S_i = \frac{n_{iY}}{n_i}
   \end{equation} 
   The above assumption that all classes are sufficiently large does not hold for high cardinality features.
   Therefore, smoothing is introduced by combining the posterior and prior probability as well as adding a weighting
   factor $\alpha$\cite{micci2001preprocessing}.
   \begin{equation}
      \label{eq:9}
      S_i = \alpha\frac{n_{iY}}{n_i} + (1-\alpha)\frac{n_Y}{n}
   \end{equation} 
   The updated equation \ref{eq:9} results in considering the overall mean $\frac{n_Y}{n}$ more strongly if the
   class $C_k$ occurs infrequently.

   \subsection*{Summary Encoder}
   Also known as the Quantile encoder, the method is aiming at incorporating quantiles instead of the mean as
   proposed with the Target encoder, in order to account for infrequent classes\cite{mougan2021quantile}.
   \begin{equation}
      \label{eq:10}
      S_i = \frac{q(x_{iY})n_i + q_p(Y)\alpha}{n_i + \alpha}
   \end{equation}
   Controlled by the number of samples per class $n_i$, the encoding technique assigns more weight to the overall
   p-quantile of the target $q_p(Y)$ if the corresponding class has infrequent occurrences.
   
   \subsection*{String Similarity}
   String similarity encoding compares class names in order to form a similarity matrix.
   While many methods exist to compare two strings with each other\cite{navigli2019overview}, the 
   Jaro-Winkler similarity\cite{cerda2018similarity} will be given as an example.
   \begin{equation}
      \label{eq:11}
      d_{jaro}(s_1,s_2) = \frac{m}{3|s_1|} + \frac{m}{3|s_2|} + \frac{m-t}{3m}  
   \end{equation}
   Eq. \ref{eq:11} shows the calculated distance based on the matching characters $m$ between a string $s_i$ and the number
   of transpositions $t$ used to account for malpositioned characters. 

   \section{Datasets}
   \noindent
   To retrieve some certainty during empirical evaluation, preprocessing steps as well as model training and
   evaluation on $10$ datasets is conducted to form a benchmark. Consideration of three data providers
   led to the conclusion that the UCI Machine Learning Repository\cite{asuncion2007uci} is used as a single
   source of data.\\
   Focusing solely on classification tasks, the following datasets for binary as well as multi-label classification
   are selected:

   \begin{Table}
      \vspace{0.25cm}
      \begin{tabular}{ |p{2.1cm}|p{1cm}|p{1.3cm}|p{1.5cm}| }
         \hline
         \multicolumn{4}{|c|}{Dataset}                                                       \\
         \hline
         Name                                         & Target & Features     & Imbalance    \\
         \hhline{|=|=|=|=|}
         Adult\cite{misc_adult_2}                     & Binary & \hfil 14     & \hfil 0.203  \\
         Mushroom\cite{misc_mushroom_73}              & Binary & \hfil 22     & \hfil 0.001  \\
         Bank\cite{misc_bank_marketing_222}           & Binary & \hfil 17     & \hfil 0.492  \\
         Breast\cite{misc_breast_cancer_wisconsin_15} & Binary & \hfil 9      & \hfil 0.070  \\
         German\cite{misc_german_credit_data_144}     & Binary & \hfil 20     & \hfil 0.118  \\
         Spambase\cite{misc_spambase_94}              & Binary & \hfil 57     & \hfil 0.032  \\
         \hline
         Car\cite{misc_car_evaluation_19}             & Multi  & \hfil 6      & \hfil 0.396  \\
         CMC\cite{misc_contraceptive_method_choice_30}& Multi  & \hfil 10     & \hfil 0.230  \\
         Nursery\cite{misc_nursery_76}                & Multi  & \hfil 8      & \hfil 0.142  \\
         Scale\cite{misc_balance_scale_12}            & Multi  & \hfil 4      & \hfil 0.343  \\
         \hline
        \end{tabular}
        \label{table:2}
        \captionof{table}{Utilized Datasets}
   \end{Table}

   \noindent
   The balance of a dataset is calculated by utilizing the Shannon Diversity Index\cite{shannon1948mathematical} in order
   to understand the label-distribution of the target.
   \begin{equation}
      \label{eq:12}
      E_H = \frac{-\sum_{i=1}^{k}\frac{n_{iY}}{n}log{(\frac{n_{iY}}{n})}}{log{(k)}}  
   \end{equation}
   Imbalance is given by $1-E_H$ where $0$ is interpreted as perfectly balanced.
   
   \section{Experiment}
   \noindent
   The first step of the experimental setup consists of preprocessing the continuous predictors in order to obtain fully
   categorical datasets. The second step describes the encoding with a predefined subset of encoding methods to cover the
   aforementioned encoder families. The last subsection focusses on the neural entity and context model, which embed and
   learn from the preprocessed data.
   
   \subsection*{Preprocessing I}
   Discretization is conducted using a decision tree model due to the advantages it offers as mentioned in the last
   paragraph of section \ref{discretization}. The procedure is described on the Adult data for exemplary purposes.\\
   Before one can start with growing a tree, the datasets need to be initially evaluated and divided into continuous and discrete
   features. Attributes with less than 10 unique values are considered discrete and will not be processed by the Discretizer.
   After partitioning, the continuous attributes are handled one at the time. Done by growing a large, 5-fold cross-validated tree and
   performing a grid search over its maximum path of seven instances from root to leave, one can obtain the cost complexity pruning path of the best estimator built.
   Using the $\alpha$-parameters of the initial tree, building a pruned version by passing one $\alpha$ at the time
   and evaluate the performance of the regularized model by conducting a second 5-fold cross-validation. It can be shown that
   pruning does affect the model's outcome only marginally and reduces the complexity to its relevant nodes as seen in figure \ref{fig:1}.
   
   \vspace{0.5cm}
   \begin{Figure}
      \includegraphics[width=\linewidth]{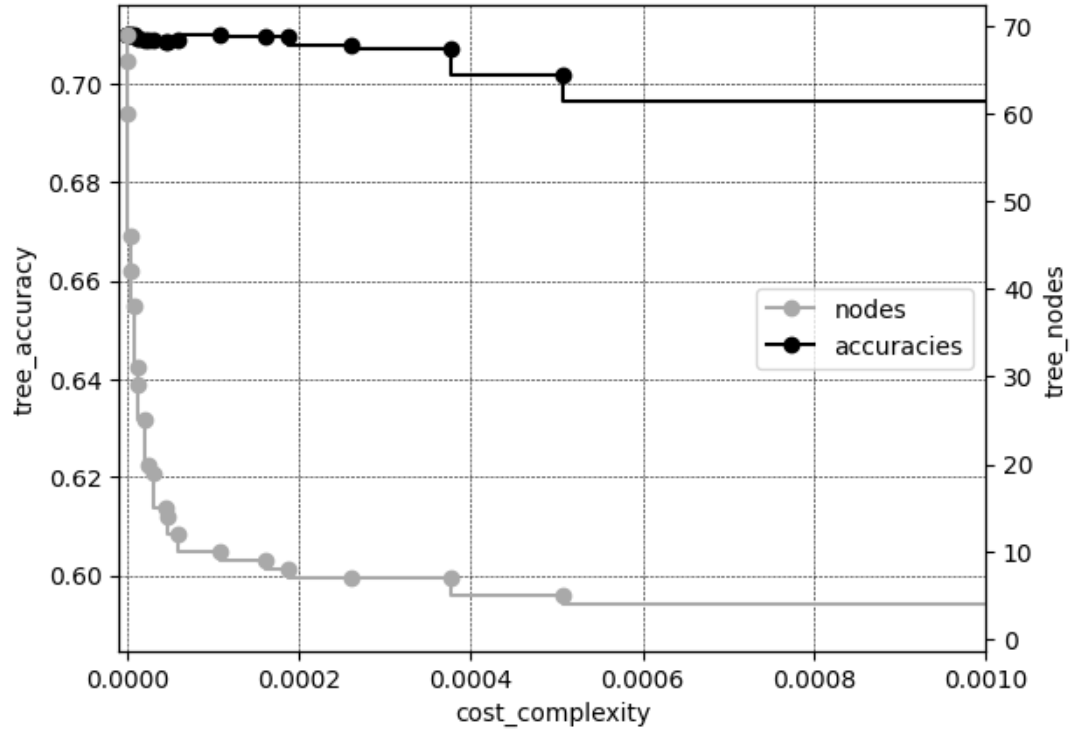}
      \captionof{figure}{Number of Decision Nodes and Accuracy}
      \label{fig:1}
   \end{Figure}
   \vspace{0.5cm}

   \noindent
   To perform discretization, the mean accuracy divided by the standard deviation of the cross-validated model is used to choose a suited
   $\alpha$. The remaining nodes represent the bin edges for interval construction, where each value of the continuous
   variable is assigned to the corresponding bin. Since building intervals from decision nodes result in a Set of
   $C_{k-1}$ classes, the first and last node is interpreted as an open interval to avoid mismatches while assigning the predictor values
   to the intervals.\\
   Using only few relevant nodes to form bins, a smoothing effect between discretized predictor and target can be observed as seen
   in figure \ref{fig:2}.
   
   \vspace{0.5cm}
   \begin{Figure}
      \includegraphics[width=\linewidth]{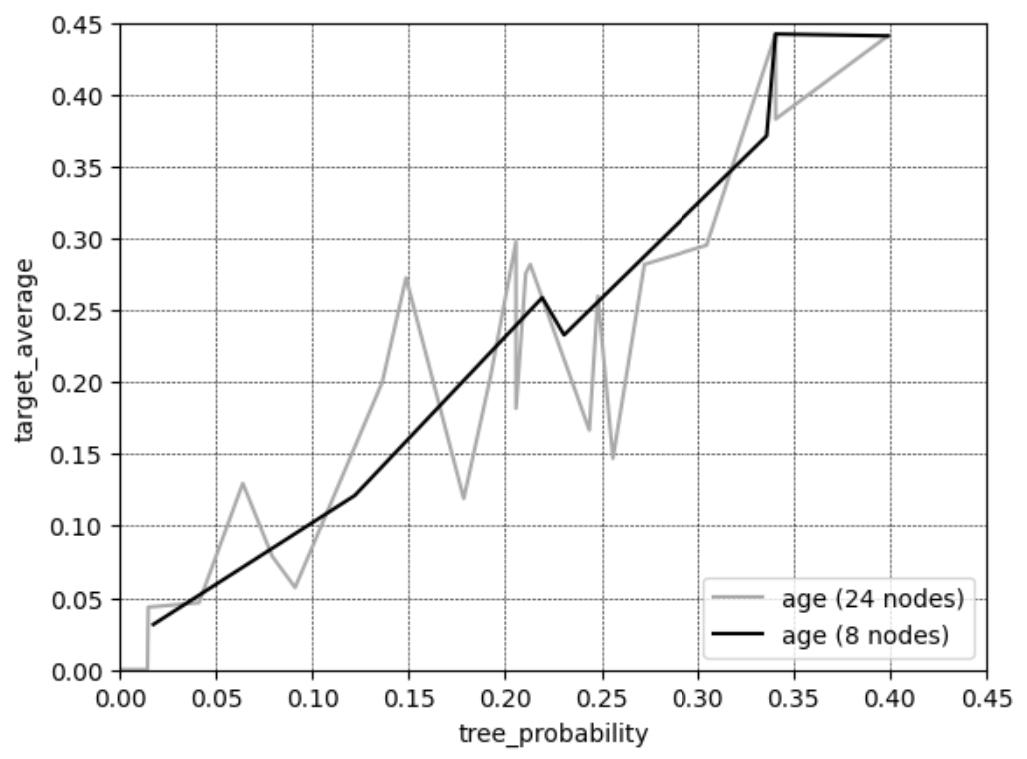}
      \captionof{figure}{Monotonic Relationship Between Predictor and Target}
      \label{fig:2}
   \end{Figure}

   \subsection*{Preprocessing II}
   The next preprocessing step includes the encoding of predictors as well as the target. Starting with the target variable
   of each dataset, the values are either Ordinal or One-Hot encoded, depending on the classification task at hand. In case of
   multi-label classification, the target additionally needs to be reshaped to a 3-dimensional format in case of the entity model,
   and a 2-dimensional format for the context model.\\
   On the other hand, six encoding techniques are applied to transform the predictor variables. Again, the Ordinal encoder is used
   to provide a baseline to compare the other methods against. Comparisons are made to the Rarelabel encoder of the same
   encoder family as well as the One-Hot, Target, Summary and String-Similarity methods. Contrast encoders were dismissed during
   evaluation due to their assumptions (e.g. levels of effect present within data or sequential dependence between instances).\\
   The transformation is applied to one feature at a time in order to handle unknown values during transformation per predictor,
   as well as obtaining a mapping for verification. The procedure is necessary due to the preceding data split into training-,
   validation- and test-set. This is achieved by wrapping the encoding methods from sklearn and the feature
   engine library. One has to be cautious to perform a train-, test- \& validation-split before fitting certain encoders to the
   data, since some of them incorporate the target variable during the procedure.

   \subsection*{Entity Model}
   Due to the architecture of the neural network, no feature scaling (e.g. normalization) is conducted during preprocessing.
   Although no statement can be made on how such scaling influences the embedding layers for tabular data, the topic has potential
   to be picked up for future investigation.\\
   The "Entity" model (base model) captivates through its simplicity. Consisting only of embedding layers
   and a multi-layer perceptron block, the network is already able to learn underlying data structures within the datasets.

   \begin{Figure}
      \vspace{0.5cm}
      \includegraphics[width=\linewidth]{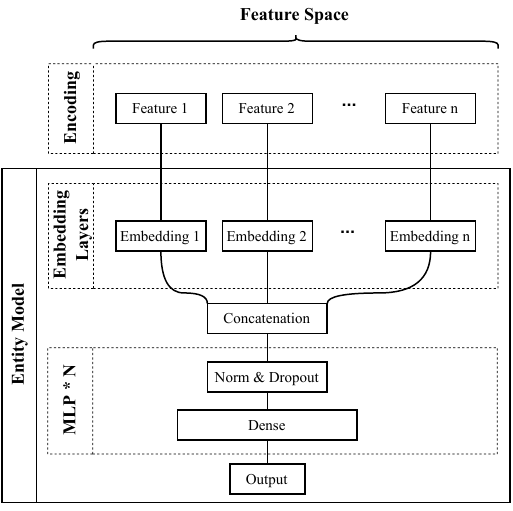}
      \captionof{figure}{Entity Model Architecture}
      \label{fig:3}
      \vspace{0.5cm}
   \end{Figure}

   \noindent
   The model itself is constructed with the functional API of the Keras library. A major distinction to other architectures can be observed
   in the parallel setup of the embedding layers. Each layer corresponds to a single feature and therefore, the number of embedding layers
   is given by the feature space the applied encoding method introduces. Furthermore, the embedding dimension is given by the number of classes $C_k$
   per predictor, resulting in a dynamic embedding space allocation\cite{lakshmanan2020machine}.
   
   \begin{equation}
      \label{eq:13}
      d = \lceil \sqrt[2]{C_k}*1.6 \rceil
   \end{equation}

   \noindent
   The embedding space $d$ is calculated by taking the square root of the number of classes in set $C$ per feature and multiplied by the constant $1.6$.
   Additionally, to obtain integer based dimensions, a ceiling function is applied.
   Although just a rule of thumb, equation \ref{eq:13} introduces suitable embedding spaces per passed feature. Smaller embeddings
   prove to be sufficient, since increasing the space often is superfluous\cite{gu2021principled}.\\
   The Input dimension for each embedding layer consists of the number of unique classes in set $C$ per feature. Additionally, $1$ is
   added to each input per layer to enable the handling of potentially unknown values during encoding.\\
   \noindent
   After concatenating all embeddings of all predictors, the vector is being fed into the multi-layer perceptron block consisting of two
   repetitions $N=2$. The size of both dense layers is dependent on the concatenated vector length. Dense layer 1 holds $50\%$ of the vector
   length as hidden units, and dense layer 2 consists of $25\%$. Added to each dense layer is normalization with $\epsilon = 1*10^{-6}$ and a dropout
   rate of $10\%$. ReLU will be applied as activation for the hidden layers of the MLP block and a Sigmoid activation function for
   the dense output layer respectively.\\
   \noindent
   The Entity model is trained using binary cross entropy as loss function and Adam\cite{kingma2014adam} as optimizer. Training is further conducted using
   $10$ epochs and a batch size of $256$ samples propagated through the network in between updates.\\
   Returning loss, accuracy and the prediction probabilities to form metrics as well as keeping track of the training time builds the
   foundation for the evaluation process. 

   \subsection*{Context Model}
   As mentioned in the previous section, the Context model shares the same foundation as the Entity model, except for the added encoding part
   of the transformer architecture\cite{vaswani2017attention}. The model is inspired by Khalid Salamas code example\footnote[1]{\url{https://keras.io/examples/structured_data/tabtransformer/}}
   found on the official Keras website and the underlying work of Xin Huang et al.\cite{Huang2020Dec}\\
   Besides the additional encoder block added to the Context model, a second change needs to be made to the embedding layers for the multi-head
   attention to work at all. Due to stacking instead of concatenating the embeddings before feeding them into the encoder block, all embedding
   layers need to have the same dimensionality.\\

   \begin{Figure}
      \vspace{0.5cm}
      \includegraphics[width=\linewidth]{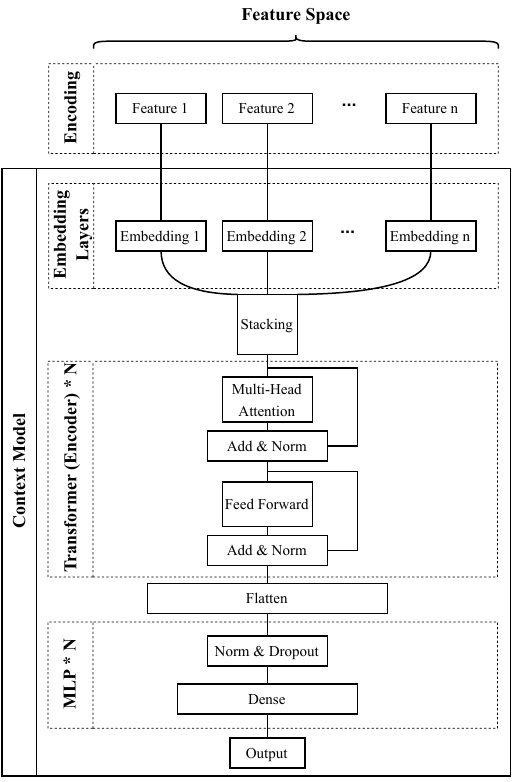}
      \captionof{figure}{Context Model Architecture}
      \label{fig:4}
      \vspace{0.5cm}
   \end{Figure}

   \noindent
   Since the embedding dimension is no longer dependent on the input feature, a fixed dimensionality of $d=10$ is applied after consideration of all datasets.\\
   Proceeding with the Transformer part, the block is implemented using only $N=1$ repetition. After multi-head attention and the feed forward
   network respectively, a skip connection and layer normalization is added. The multi-head comprises 4 attention heads, the embedding dimensionality
   $d=10$ as well as a dropout rate of $10$ Percent.\\
   Implementing the feed forward network, the MLP block, consisting only of $N=1$ repetition, is reused. No reduction is performed with
   regard to the hidden units of the dense layer, as described during the MLP setup.\\

   \noindent
   The model run is conducted using the same hyperparameters as for the base model. A slight change is made with regard to the batch size, 
   reducing the number of samples from previously $256$ to $128$. Arguably, the training length might be increased in order to check for further
   potential of the context embeddings. Due to comparability of the two models, it was decided to leave the number of epochs identical. 

   \section{Results}
   \noindent
   The experiment was conducted five times while keeping discretization as well as train-, test- and validation-split unchanged for model training.
   The procedure allowed for variation due to randomly initialized embeddings, resulting in slightly deviating outcomes.
   Furthermore, having fully categorical datasets reduced the result set drastically, which made it possible to compute $120$ models on an Intel Core
   i7-9750H within approx-

\end{multicols}

\noindent
\begin{table}[H]
\vspace{0.25cm}
\begin{tabular}{|p{1.7cm}|p{1.8cm}|p{1.8cm} p{1.8cm} p{1.8cm} p{1.8cm} p{1.8cm}|}
   \hline
   \multicolumn{7}{|c|}{Entity Model}  \\
   \hline
   Dataset     & Ordinal               & One-Hot                     & Rarelabel                   & String Sim.                 & Summary                     & Target                \\
   \hhline{|=|=|=|=|=|=|=|}
   Adult       & $0.71 \:\:\; (0.01)$  & $0.69 \:\:\; (0.03)$        & $0.64 \:\:\; (0.02)$        & $\pmb{0.72} \:\:\; (0.02)$  & \textit{undefined}          & $0.04 \:\:\; (0.0)$   \\
   \hline
   Mushroom    & $1.0 \quad\: (0.0)$   & $1.0 \quad\: (0.0)$         & $1.0 \quad\: (0.0)$         & $1.0 \quad\: (0.0)$         & $0.99 \:\:\; (0.0)$         & $0.97 \:\:\; (0.0)$   \\
   \hline            
   Bank        & $0.55 \:\:\; (0.02)$  & $0.54 \:\:\; (0.02)$        & $0.53 \:\:\; (0.01)$        & $0.53 \:\:\; (0.04)$        & \textit{undefined}          & \textit{undefined}    \\
   \hline
   Breast      & $0.98 \:\:\; (0.02)$  & $0.98 \:\:\; (0.03)$        & $\pmb{0.99} \:\:\; (0.02)$  & $0.98 \:\:\; (0.03)$        & $\pmb{0.99} \:\:\; (0.02)$  & \textit{undefined}    \\
   \hline
   German      & $0.49 \:\:\; (0.05)$  & $0.49 \:\:\; (0.04)$        & $0.45 \:\:\; (0.06)$        & $\pmb{0.51} \:\:\; (0.03)$  & $0.43 \:\:\; (0.05)$        & \textit{undefined}    \\
   \hline
   Spambase    & $0.97 \:\:\; (0.01)$  & $0.77 \:\:\; (0.43)$        & $0.97 \:\:\; (0.01)$        & $0.97 \:\:\; (0.0)$         & $0.93 \:\:\; (0.0)$         & \textit{undefined}    \\
   \hline
   Car         & $0.79 \:\:\; (0.04)$  & $\pmb{0.91} \:\:\; (0.02)$  & $\pmb{0.81} \:\:\; (0.05)$  & $\pmb{0.88} \:\:\; (0.04)$  & $0.70 \:\:\; (0.01)$        & $0.72 \:\:\; (0.02)$  \\
   \hline
   CMC         & $0.37 \:\:\; (0.07)$  & $\pmb{0.42} \:\:\; (0.06)$  & $\pmb{0.40} \:\:\; (0.04)$  & $\pmb{0.42} \:\:\; (0.03)$  & $\pmb{0.46} \:\:\; (0.05)$  & $0.20 \:\:\; (0.12)$  \\
   \hline
   Nursery     & $0.97 \:\:\; (0.01)$  & $\pmb{1.0} \quad\: (0.0)$   & $\pmb{0.98} \:\:\; (0.0)$   & $\pmb{1.0} \quad\: (0.0)$   & $0.82 \:\:\; (0.0)$         & $0.81 \:\:\; (0.0)$   \\
   \hline
   Scale       & $0.54 \:\:\; (0.19)$  & $\pmb{0.87} \:\:\; (0.06)$  & $\pmb{0.57} \:\:\; (0.19)$  & $\pmb{0.88} \:\:\; (0.02)$  & $\pmb{0.56} \:\:\; (0.17)$  & $0.18 \:\:\; (0.25)$  \\
   \hline
  \end{tabular}
  \label{table:3}
  {\raggedright\textit{Note}: Mean and Standard Deviation, where undefined denotes absence of true positive instances.\par}
  \captionof{table}{Entity Model: F1-Score per Dataset and Encoding Technique}
\end{table}

\begin{multicols}{2}
   \noindent
   imately $4$ hours per experimental run.\\
   Next to the binary cross entropy loss, the F1-Score\cite{derczynski2016complementarity} was computed to compare the model
   outputs with different encoding methods against each other.
   Interpretation is done purely on the F-measure since it incorporates precision and recall and therefore holds additional information if the built models were able
   learn from the encoded data. Computation of the loss is found in the appendix.\\
   Comparison is made between the Ordinal encoder as baseline and the remaining encoding methods. Bold numbers represent a better result as the
   baseline. Additionally, some methods returned an undefined result due to the number of true positive instances being $0$.\\
   \noindent
   String Similarity encoding worked outstandingly well, performing better than Ordinal encoding on $6$ datasets and equally or better as the
   baseline $9$ out of $10$ times. One-Hot, Rarelabel and String Similarity also outperformed the baseline on multi-label classification
   problems without exception. On the Car and Scale datasets were improvements up to 15 and 63 percent observed.\\
   Although the context model had the same number of epochs for training as the entity model, improvement on Ordinal encoding is noted
   within the multi-label classification tasks. Despite the gain of the baseline, the One-Hot and String Similarity methods are still able to outperform Ordinal
   encoding on the context embeddings.
   Overall, String Similarity encoding achieved the same or improved results on $7$ out of $9$ datasets compared to the baseline.\\
   Results suggest that using the String Similarity method for classification yields better outcomes 

\end{multicols}

\begin{table}[H]
\vspace{0cm}
\begin{tabular}{|p{1.7cm}|p{1.8cm}|p{1.8cm} p{1.8cm} p{1.8cm} p{1.8cm} p{1.8cm}|}
   \hline
   \multicolumn{7}{|c|}{Context Model}  \\
   \hline
   Dataset     & Ordinal               & One-Hot                     & Rarelabel                   & String Sim.                 & Summary                     & Target                \\
   \hhline{|=|=|=|=|=|=|=|}
   Adult       & $0.71 \:\:\; (0.01)$  & $0.69 \:\:\; (0.02)$        & $0.63 \:\:\; (0.01)$        & $0.69 \:\:\; (0.03)$        & \textit{undefined}          & $0.04 \:\:\; (0.0)$   \\
   \hline
   Mushroom    & $1.0 \quad\: (0.0)$   & $1.0 \quad\: (0.0)$         & $1.0 \quad\: (0.0)$         & $1.0 \quad\: (0.0)$         & $0.99 \:\:\; (0.0)$         & $0.97 \:\:\; (0.0)$   \\
   \hline            
   Bank        & $0.54 \:\:\; (0.03)$  & $0.54 \:\:\; (0.04)$        & $0.51 \:\:\; (0.02)$        & $\pmb{0.55} \:\:\; (0.03)$  & \textit{undefined}          & \textit{undefined}    \\
   \hline
   Breast      & $0.99 \:\:\; (0.02)$  & $0.96 \:\:\; (0.04)$        & $0.98 \:\:\; (0.03)$        & $0.95 \:\:\; (0.03)$        & $0.96 \:\:\; (0.0)$         & \textit{undefined}    \\
   \hline
   German      & $0.48 \:\:\; (0.05)$  & $0.44 \:\:\; (0.06)$        & $0.42 \:\:\; (0.04)$        & $0.45 \:\:\; (0.06)$        & $0.41 \:\:\; (0.09)$        & \textit{undefined}    \\
   \hline
   Spambase    & $0.96 \:\:\; (0.01)$  & $0.95 \:\:\; (0.02)$        & $\pmb{0.97} \:\:\; (0.01)$  & $0.96 \:\:\; (0.01)$        & $0.92 \:\:\; (0.01)$        & \textit{undefined}    \\
   \hline
   Car         & $0.86 \:\:\; (0.01)$  & $\pmb{0.93} \:\:\; (0.02)$  & $0.85 \:\:\; (0.01)$        & $\pmb{0.92} \:\:\; (0.01)$  & $0.70 \:\:\; (0.01)$        & $0.69 \:\:\; (0.01)$  \\
   \hline
   CMC         & $0.37 \:\:\; (0.11)$  & $\pmb{0.41} \:\:\; (0.04)$  & $\pmb{0.40} \:\:\; (0.05)$  & $\pmb{0.41} \:\:\; (0.03)$  & $\pmb{0.47} \:\:\; (0.03)$  & $0.26 \:\:\; (0.02)$  \\
   \hline
   Nursery     & $0.99 \:\:\; (0.01)$  & $\pmb{1.0} \quad\: (0.0)$   & $0.98 \:\:\; (0.01)$        & $\pmb{1.0} \quad\: (0.0)$   & $0.82 \:\:\; (0.0)$         & $0.82 \:\:\; (0.01)$  \\
   \hline
   Scale       & $0.74 \:\:\; (0.19)$  & $\pmb{0.92} \:\:\; (0.02)$  & $0.54 \:\:\; (0.36)$        & $\pmb{0.92} \:\:\; (0.02)$  & $0.64 \:\:\; (0.0)$         & $0.22 \:\:\; (0.31)$  \\
   \hline
  \end{tabular}
  \label{table:4}
  {\raggedright \textit{Note}: Mean and Standard Deviation, where undefined denotes absence of true positive instances.\par}
  \captionof{table}{Context Model: F1-Score per Dataset and Encoding Technique}
\end{table}

\begin{figure}[ht]%
   \centering
   \subfloat[\centering Mean Training Time Entity Model]{{\includegraphics[width=7.5cm]{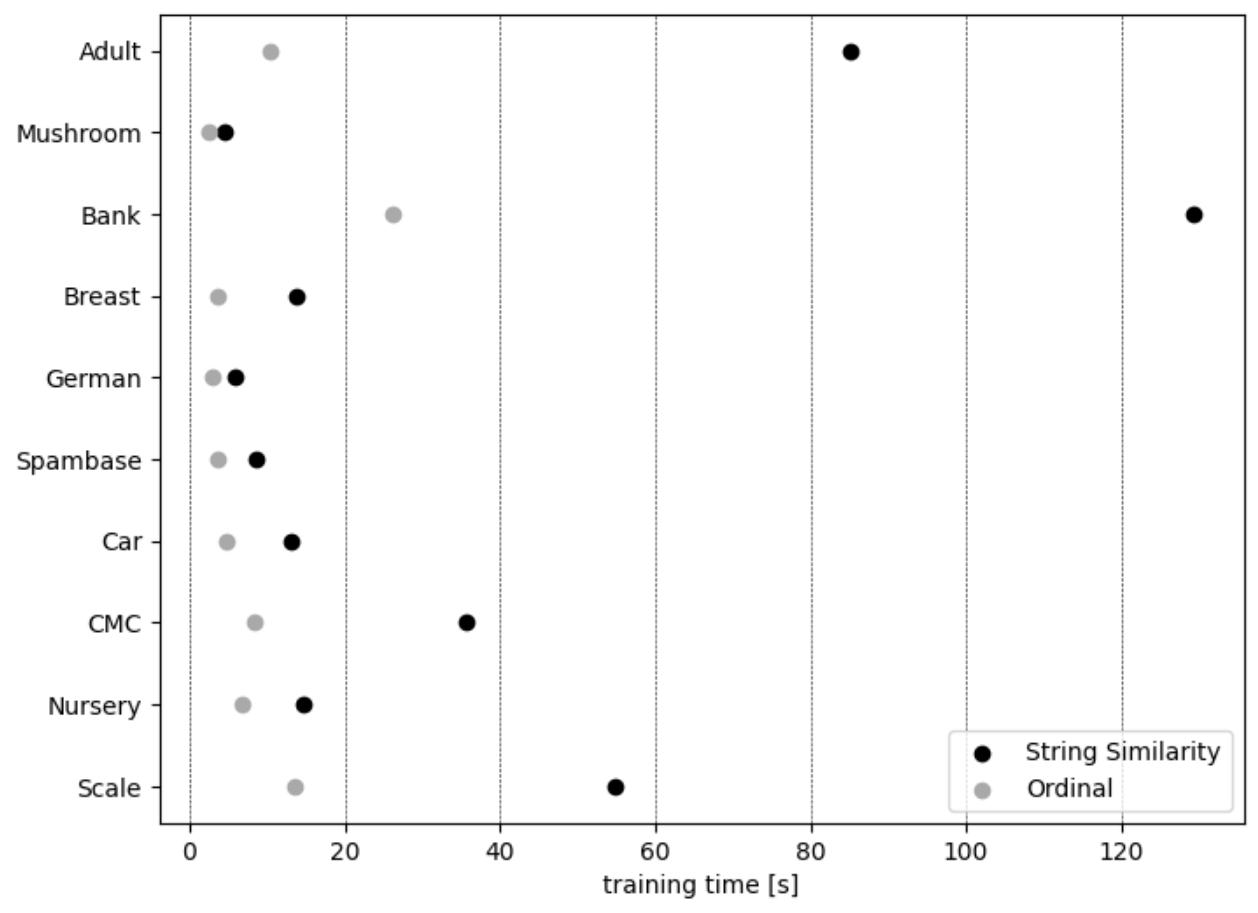} }}%
   \qquad
   \subfloat[\centering Mean Training Time Context Model]{{\includegraphics[width=6.65cm]{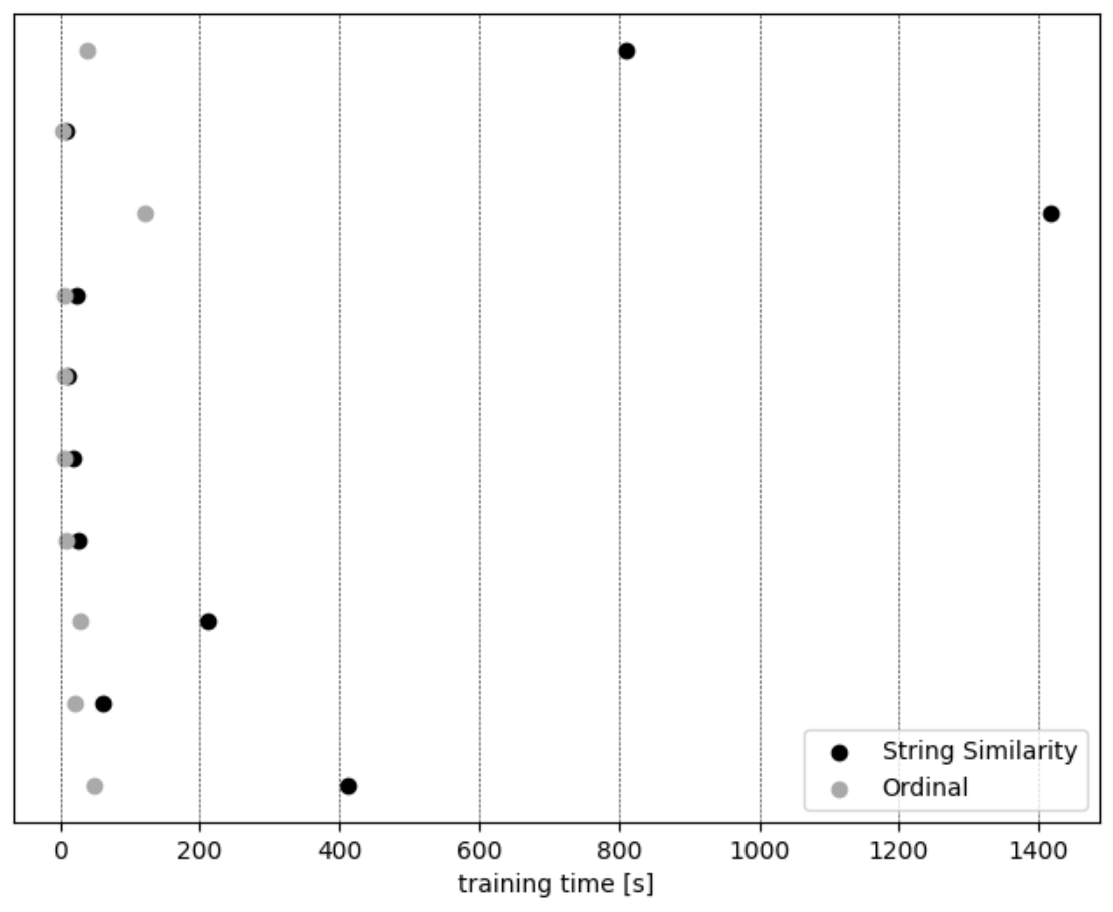} }}%
   \caption{Training Times Ordinal versus String Similarity Encoding per Dataset}%
   \label{fig:5}%
\end{figure}

\begin{multicols}{2}
   \noindent
   than using the standard Ordinal encoding technique. Nonetheless, testing different methods for a specific use case is always advised.\\
   Achieving better performance usually comes with some sort of cost. In this case, the major drawback lies within computation time. Especially
   high cardinality predictors are expensive due to rapid expansion of the feature space.
   The behaviour can be observed when comparing the mean training time of Ordinal Encoding to the String Similarity method for both models.

   \section{Future Work}
   \noindent
   During exploration as well as experimental setup, further questions occurred which make interesting topics for potential future
   studies:\\
   \noindent
   First and foremost, no comparison was made regarding neural networks with and without fully discretized datasets. The standard procedure
   consists of implementing two inputs, one for continuous and another for discrete variables. It is unclear how different encoding methods
   affect architectures with two inputs.\\
   Second, no further evaluation was made regarding the performance of a single encoding technique. Why does it seem more difficult for neural
   networks to learn from target encoded data? Does the behaviour hold true for other encoders of the target encoder family? It might be just convenient
   for the model to learn the incorporated mean/quantiles of the target instead of less obvious data structures. The behaviour certainly needs to be
   further assessed in order to gain more understanding.\\
   Lastly, One-Hot encoding destroys the class structure of the predictor by introducing a set $X$ of new binary features. Although this is the case, the
   models seemingly benefited from the method. It could be interesting to investigate how the encoders affect such class structures by 
   analysis of the entity and context embeddings.

\end{multicols}


\printbibliography


\section*{Supplements}

\begin{table}[H]
   \begin{tabular}{|p{1.7cm}|p{1.8cm}|p{1.8cm} p{1.8cm} p{1.8cm} p{1.8cm} p{1.8cm}|}
      \hline
      \multicolumn{7}{|c|}{Entity Model}  \\
      \hline
      Dataset     & Ordinal            & One-Hot            & Rarelabel          & String Sim.        & Summary            & Target             \\
      \hhline{|=|=|=|=|=|=|=|}
      Adult       & $3.033\text{e-}1$  & $2.997\text{e-}1$  & $3.56\text{e-}1$   & $2.998\text{e-}1$  & $5.858\text{e-}1$  & $5.829\text{e-}1$  \\
      \hline
      Mushroom    & $5.6\text{e-}5$    & $9.6\text{e-}5$    & $5.8\text{e-}5$    & $1.01\text{e-}4$   & $4.726\text{e-}3$  & $1.197\text{e-}1$  \\
      \hline            
      Bank        & $1.752\text{e-}1$  & $1.742\text{e-}1$  & $1.811\text{e-}1$  & $1.737\text{e-}1$  & $3.277\text{e-}1$  & $3.276\text{e-}1$  \\
      \hline
      Breast      & $7.225\text{e-}2$  & $4.297\text{e-}2$  & $7.168\text{e-}2$  & $5.274\text{e-}2$  & $4.291\text{e-}2$  & $6.914\text{e-}1$  \\
      \hline
      German      & $4.257\text{e-}1$  & $4.621\text{e-}1$  & $4.545\text{e-}1$  & $4.578\text{e-}1$  & $4.685\text{e-}1$  & $5.137\text{e-}1$  \\
      \hline
      Spambase    & $8.461\text{e-}2$  & $1.352$            & $7.953\text{e-}2$  & $9.668\text{e-}2$  & $1.656\text{e-}1$  & $6.822\text{e-}1$  \\
      \hline
      Car         & $2.606\text{e-}1$  & $1.235\text{e-}1$  & $2.653\text{e-}1$  & $1.402\text{e-}1$  & $2.633\text{e-}1$  & $2.843\text{e-}1$  \\
      \hline
      CMC         & $5.941\text{e-}1$  & $5.764\text{e-}1$  & $5.971\text{e-}1$  & $5.779\text{e-}1$  & $5.47\text{e-}1$   & $6.484\text{e-}1$  \\
      \hline
      Nursery     & $3.508\text{e-}2$  & $6.282\text{e-}3$  & $2.581\text{e-}2$  & $4.309\text{e-}3$  & $1.942\text{e-}1$  & $1.933\text{e-}1$  \\
      \hline
      Scale       & $5.298\text{e-}1$  & $3.2\text{e-}1$    & $5.407\text{e-}1$  & $2.911\text{e-}1$  & $5.049\text{e-}1$  & $5.762\text{e-}1$  \\
      \hline
     \end{tabular}
     \vspace{0.1cm}
     \label{table:5}
     \captionof{table}{Entity Model: Mean Binary Cross Entropy Loss per Dataset and Encoding Technique}
   \end{table}

   \begin{table}[H]
      \begin{tabular}{|p{1.7cm}|p{1.8cm}|p{1.8cm} p{1.8cm} p{1.8cm} p{1.8cm} p{1.8cm}|}
         \hline
         \multicolumn{7}{|c|}{Entity Model}  \\
         \hline
         Dataset     & Ordinal            & One-Hot            & Rarelabel          & String Sim.        & Summary            & Target             \\
         \hhline{|=|=|=|=|=|=|=|}
         Adult       & $2.226\text{e-}3$  & $1.102\text{e-}2$  & $7.73\text{e-}3$   & $1.624\text{e-}3$  & $1.51\text{e-}3$   & $3.013\text{e-}3$  \\
         \hline
         Mushroom    & $1.2\text{e-}5$    & $4.2\text{e-}5$    & $2.3\text{e-}5$    & $4.5\text{e-}5$    & $7.238\text{e-}3$  & $4.167\text{e-}3$  \\
         \hline            
         Bank        & $1.261\text{e-}3$  & $3.081\text{e-}3$  & $4.148\text{e-}3$  & $2.843\text{e-}3$  & $1.576\text{e-}3$  & $9.7\text{e-}4$    \\
         \hline
         Breast      & $1.282\text{e-}2$  & $1.519\text{e-}2$  & $1.807\text{e-}2$  & $2.902\text{e-}2$  & $2.9\text{e-}2$    & $1.455\text{e-}2$  \\
         \hline
         German      & $2.989\text{e-}2$  & $2.179\text{e-}2$  & $2.179\text{e-}2$  & $2.797\text{e-}2$  & $1.747\text{e-}2$  & $1.704\text{e-}2$  \\
         \hline
         Spambase    & $6.116\text{e-}3$  & $2.806$            & $8.749\text{e-}3$  & $2.494\text{e-}3$  & $8.625\text{e-}3$  & $4.461\text{e-}3$  \\
         \hline
         Car         & $3.628\text{e-}2$  & $1.049\text{e-}2$  & $7.573\text{e-}2$  & $2.255\text{e-}2$  & $8.029\text{e-}3$  & $4.437\text{e-}2$  \\
         \hline
         CMC         & $8.619\text{e-}3$  & $2.715\text{e-}3$  & $1.649\text{e-}2$  & $3.848\text{e-}3$  & $8.844\text{e-}3$  & $1.787\text{e-}3$  \\
         \hline
         Nursery     & $8.684\text{e-}3$  & $3.353\text{e-}3$  & $2.768\text{e-}3$  & $4.25\text{e-}4$   & $9.28\text{e-}4$   & $8.39\text{e-}4$   \\
         \hline
         Scale       & $6.325\text{e-}2$  & $8.197\text{e-}2$  & $9.397\text{e-}2$  & $2.335\text{e-}2$  & $2.663\text{e-}2$  & $2.031\text{e-}2$  \\
         \hline
        \end{tabular}
        \vspace{0.1cm}
        \label{table:6}
        \captionof{table}{Entity Model: Standard Deviation Binary Cross Entropy Loss per Dataset and Encoding Technique}
      \end{table}

   \begin{table}[H]
      \begin{tabular}{|p{1.7cm}|p{1.8cm}|p{1.8cm} p{1.8cm} p{1.8cm} p{1.8cm} p{1.8cm}|}
         \hline
         \multicolumn{7}{|c|}{Context Model}  \\
         \hline
         Dataset     & Ordinal            & One-Hot            & Rarelabel          & String Sim.        & Summary            & Target             \\
         \hhline{|=|=|=|=|=|=|=|}
         Adult       & $3.011\text{e-}1$  & $3.011\text{e-}1$  & $3.588\text{e-}1$  & $3.044\text{e-}1$  & $5.892\text{e-}1$  & $5.798\text{e-}1$  \\
         \hline
         Mushroom    & $9\text{e-}6$      & $2\text{e-}6$      & $1.4\text{e-}5$    & $3\text{e-}6$      & $2.987\text{e-}3$  & $1.173\text{e-}1$  \\
         \hline            
         Bank        & $1.761\text{e-}1$  & $1.771\text{e-}1$  & $1.83\text{e-}1$   & $1.776\text{e-}1$  & $3.265\text{e-}1$  & $3.272\text{e-}1$  \\
         \hline
         Breast      & $3.27\text{e-}2$   & $5.392\text{e-}2$  & $3.351\text{e-}2$  & $5.816\text{e-}2$  & $5.423\text{e-}2$  & $6.781\text{e-}1$  \\
         \hline
         German      & $4.321\text{e-}1$  & $4.822\text{e-}1$  & $4.561\text{e-}1$  & $4.721\text{e-}1$  & $4.63\text{e-}1$   & $5.155\text{e-}1$  \\
         \hline
         Spambase    & $9.358\text{e-}2$  & $1.224\text{e-}1$  & $8.407\text{e-}2$  & $1.044\text{e-}1$  & $1.688\text{e-}1$  & $6.815\text{e-}1$  \\
         \hline
         Car         & $1.626\text{e-}1$  & $8.907\text{e-}2$  & $1.708\text{e-}1$  & $9.005\text{e-}2$  & $2.437\text{e-}1$  & $2.374\text{e-}1$  \\
         \hline
         CMC         & $5.993\text{e-}1$  & $5.726\text{e-}1$  & $5.911\text{e-}1$  & $5.851\text{e-}1$  & $5.334\text{e-}1$  & $6.548\text{e-}1$  \\
         \hline
         Nursery     & $1.622\text{e-}2$  & $2.1\text{e-}3$    & $2.277\text{e-}2$  & $1.88\text{e-}3$   & $1.927\text{e-}1$  & $1.931\text{e-}1$  \\
         \hline
         Scale       & $4.428\text{e-}1$  & $1.904\text{e-}1$  & $4.876\text{e-}1$  & $1.862\text{e-}1$  & $4.812\text{e-}1$  & $6.124\text{e-}1$  \\
         \hline
        \end{tabular}
        \vspace{0.1cm}
        \label{table:7}
        \captionof{table}{Context Model: Mean Binary Cross Entropy Loss per Dataset and Encoding Technique}
      \end{table}

      \begin{table}[H]
         \begin{tabular}{|p{1.7cm}|p{1.8cm}|p{1.8cm} p{1.8cm} p{1.8cm} p{1.8cm} p{1.8cm}|}
            \hline
            \multicolumn{7}{|c|}{Context Model}  \\
            \hline
            Dataset     & Ordinal            & One-Hot            & Rarelabel          & String Sim.        & Summary            & Target             \\
            \hhline{|=|=|=|=|=|=|=|}
            Adult       & $5.803\text{e-}3$  & $8.223\text{e-}3$  & $4.62\text{e-}3$   & $5.822\text{e-}3$  & $3.079\text{e-}3$  & $2.903\text{e-}3$  \\
            \hline
            Mushroom    & $3\text{e-}5$      & $1\text{e-}5$      & $4\text{e-}5$      & $1\text{e-}5$      & $1.832\text{e-}3$  & $8.07\text{e-}4$   \\
            \hline            
            Bank        & $3.906\text{e-}3$  & $5.329\text{e-}3$  & $2.417\text{e-}3$  & $3.899\text{e-}3$  & $5.09\text{e-}4$   & $8.64\text{e-}4$   \\
            \hline
            Breast      & $2.995\text{e-}3$  & $2.634\text{e-}2$  & $3.398\text{e-}2$  & $2.181\text{e-}2$  & $1.592\text{e-}2$  & $5.117\text{e-}3$  \\
            \hline
            German      & $3.132\text{e-}2$  & $5.755\text{e-}2$  & $1.711\text{e-}2$  & $5.022\text{e-}2$  & $2.614\text{e-}2$  & $1.187\text{e-}2$  \\
            \hline
            Spambase    & $1.35\text{e-}2$   & $3.167\text{e-}2$  & $1.256\text{e-}2$  & $7.383\text{e-}3$  & $1.477\text{e-}2$  & $1.084\text{e-}3$  \\
            \hline
            Car         & $1.361\text{e-}2$  & $1.43\text{e-}2$   & $4.048\text{e-}2$  & $1.174\text{e-}2$  & $7.681\text{e-}3$  & $4.163\text{e-}3$  \\
            \hline
            CMC         & $1.618\text{e-}2$  & $1.352\text{e-}2$  & $1.578\text{e-}2$  & $9.084\text{e-}3$  & $9.512\text{e-}3$  & $5.141\text{e-}3$  \\
            \hline
            Nursery     & $4.756\text{e-}3$  & $6.43\text{e-}4$   & $6.772\text{e-}3$  & $6.77\text{e-}4$   & $1.104\text{e-}3$  & $9.61\text{e-}4$   \\
            \hline
            Scale       & $1.019\text{e-}1$  & $1.13\text{e-}2$   & $8.888\text{e-}2$  & $1.969\text{e-}2$  & $8.813\text{e-}3$  & $4.977\text{e-}2$  \\
            \hline
           \end{tabular}
           \vspace{0.1cm}
           \label{table:8}
           \captionof{table}{Context Model: Standard Deviation Binary Cross Entropy Loss per Dataset and Encoding Technique}
         \end{table}

\end{document}